\documentclass{article}
\usepackage{ijcai26}

\usepackage{times}
\usepackage{soul}
\usepackage{url}
\usepackage[hidelinks]{hyperref}
\usepackage[utf8]{inputenc}
\usepackage[small]{caption}
\usepackage{graphicx}
\usepackage{amsmath}
\usepackage{amsthm}
\usepackage{amssymb}
\usepackage{booktabs}
\usepackage{pifont}
\usepackage{enumitem}
\usepackage[ruled,vlined]{algorithm2e}
\usepackage{float}
\usepackage{siunitx} 
\usepackage[switch]{lineno}
\usepackage{mdframed}
\usepackage{xcolor}
\usepackage{array}
\usepackage{geometry} 
\usepackage{multirow}
\newtheorem{example}{Example}
\newtheorem{theorem}{Theorem}
\newtheorem{definition}{Definition}
\newtheorem{prop}{Proposition}

\newtheorem{corollary}{Corollary}
\newcommand{\defeq}{\triangleq}

\title{Abstraction Generation for Generalized Planning with Pretrained Large Language Models}

\author{
    Author Name
    \affiliations
    Affiliation
    \emails
    email@example.com
}

\author{
Zhenhe Cui$^{1}$
\and
Huaxiang Xia$^1$\and
Hangjun Shen$^{1}$\and
Kailun Luo$^{2}$\footnote{Corresponding author}\and
Yong He$^1$\and
Wei Liang$^1$
\\
\affiliations
$^1$Hunan University of Science and Technology, Xiangtan 411201, China\\
$^2$Dongguan University of Technology, Dongguan 523808, China\\
\emails
zhhcui@hnust.edu.cn,
huaxiangxia@outlook.com,
h70094650@gmail.com,
luokl@dgut.edu.cn,
yonghe@hnust.edu.cn,
wliang@hnust.edu.cn
}

\begin{document}

\maketitle

\begin{abstract}
Qualitative Numerical Planning (QNP) serves as an important abstraction model for generalized planning (GP), which aims to compute general plans that solve multiple instances at once. Recent works show that large language models (LLMs) can function as generalized planners. This work investigates whether LLMs can serve as QNP abstraction generators for GP problems and how to fix abstractions via automated debugging. We propose a prompt protocol: input a GP domain and training tasks to LLMs, prompting them to generate abstract features and further abstract the initial state, action set, and goal into QNP problems. An automated debugging method is designed to detect abstraction errors, guiding LLMs to fix abstractions. Experiments demonstrate that
under properly guided by automated debugging, some LLMs can generate useful QNP abstractions. 
\end{abstract}

\begin{figure*}[t]
\centering
\includegraphics[width=\textwidth]{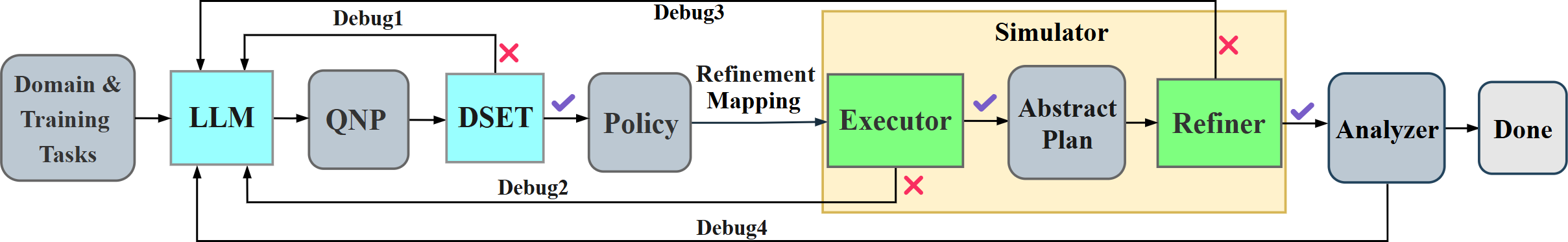}
\caption{The overall framework for abstraction generation for generalized planning with pretrained LLMs.}
\label{fig:system_arch}
\end{figure*}

\section{Introduction}

Abstraction is an important method for GP,  which aims at synthesizing a single plan works for multiple planning instances \cite{bonet2009automatic,2016Generalized,2011Qualitative,2018Features,2019Illanes,Cui21}. The idea of abstraction is to develop an abstract model for the problem that suppresses less important details, find a solution in the abstract model, and use this solution to guide the search for a solution in the concrete problem. The widely used abstract model QNP (qualitative numerical planning) for GP is introduced by \cite{2011Qualitative}. QNPs are classical planning problems extended with non-negative real variables that can be increased or decreased by some arbitrary amount. 
\citeauthor{2018Features} \shortcite{2018Features} proposed solving a GP problem by abstracting it into a QNP problem, they showed that if the abstraction is sound, then a solution to the QNP is also a solution to the original problem.  \citeauthor{Cui21} \shortcite{Cui21}  proposed an abstraction framework for GP based on situation calculus and Golog. Relating a QNP problem to a GP problem by the notion of a refinement mapping, they gave definitions of sound/complete QNP abstractions of a GP problem, and showed that solutions to a GP problem are nicely related to those of its sound/complete QNP abstractions. In particular, the refinement of any solution to a sound QNP abstraction is a solution to the original GP problem.

Recently, LLMs have shown promise for supporting GP. \citeauthor{SilverDSTK024}~\shortcite{SilverDSTK024} showed that GPT-4 can act as a generalized planner. 
They integrate Chain-of-Thought summarization and automated debugging with four types of feedback, generating domain-specific generalized plan from a small set of training tasks. Their work demonstrates GPT-4’s strong performance in GP. However, it remains unclear whether similar success can be achieved for \emph{abstraction} generation. 
\citeauthor{banihashemi2025using}~\shortcite{banihashemi2025using} study the use of LLMs to produce abstractions for individual planning instances. However, how to generate a \emph{generalized abstraction} that works across multiple instances, and how to detect to correct abstraction errors, remains open.



In this work, we ask the question: \textit{can pretrained LLMs serve as GP abstraction generators for PDDL domains?} Specifically, we investigate whether existing LLMs can be used to generate QNP abstractions for GP, and how to prompt them to revise abstractions in case of abstraction errors. For each GP problem, we input the planning domain and some training instances to the LLM. We first prompt LLM to generate abstract features according to predefined feature templates. The obtained features are served as the state variables for a QNP abstraction. We then prompt the LLM to abstract the initial state, action set, and goal of the given training instances based on the generated features, and finally output the resulting QNP abstraction.


To improve the applicability of LLM-generated QNP abstractions, we propose an automated debugging method based on the notion of approximate abstraction  introduced by \citeauthor{Bonet19}~\shortcite{Bonet19}.
We preform the following process: for a QNP abstraction generated by the LLM, we first use the QNP solver DSET \cite{ZengLL22} to obtain an abstract solution; 
we then refine the abstract solution into an action sequence for the given training instances; we finally check whether there exists an action sequence achieving the goal of the training instances. The above process might fails, and during the process, we continuously track abstraction errors and prompt the LLM to revise the abstraction based on these error information. We repeat this automated debugging process until a predetermined number of iterations is reached or a usable abstraction is obtained.


We evaluated our approach on seven GP domains. We considered four LLMs: GPT, Gemini, DeepSeek, and Qwen, both with and without our debugging process. The results indicate that our method can generate useful QNP abstractions in most domains. We show that some LLMs can serve as useful QNP abstraction generators for GP, if a guided debugging process is provided.


\section{Preliminaries}

\subsection{Generalized Planning}

We use the FOL with counting (FOC)  \cite{Kuske2017First} as the underlying logic.

A GP domain is a tuple $\mathcal{D}=\langle \mathcal{P}, \mathcal{O}, \mathcal{A}\rangle$, where $\mathcal{P}$ is a set of predicates, $\mathcal{O}$ is a  set of objects and $\mathcal{A}$ is a set of action schemata. Each $a(\vec{x})\in \mathcal{A}$ consists of precondition $\mathtt{pre}(a(\vec{x}))$ and effect $\mathtt{eff}(a(\vec{x}))$,
where $\mathtt{pre}(a(\vec{x}))$ is a set of first-order literals defined over $\mathcal{P}$ and $\mathcal{O}$, and $\mathtt{eff}(a(\vec{x}))$ consists of add list $\mathtt{add}(\vec{x})$ and delete list $\mathtt{del}(\vec{x})$, both of which are first-order atoms defined on $\mathcal{P}$ and $\mathcal{O}$.




A GP problem is a tuple $\mathcal{Q}=\langle \mathcal{D},  S_0,  S_G\rangle$, where $\mathcal{D}$ is a GP domain, $S_0$ is the initial condition, and $S_G$ is the goal condition. Both $S_0$ and $S_G$ are FOL formulae over predicates in $\mathcal{P}$ and objects in $\mathcal{O}$. A GP instance is a tuple $Q=\langle \mathcal{D},  s_0, s_g\rangle$, where $s_0$ and $s_g$ are states satisfying $S_0$ and $S_G$. 
A state $s$ of $Q$ means a set of literals defined over predicates in $\mathcal{P}$ and objects in $\mathcal{O}$.



Given an action schema $a(\vec{x})$ and a sequence of objects $\vec{o}$, a ground action $a(\vec{o})$ (or $a$ for short) is the result of replacing each occurrence of each variable $x_{i}$ in $\vec{x}_{i}$ with object $o_{i}$ in $\vec{o}_{i}$.
We say that $a(\vec{o})$ is applicable in $s$ if $s\models \mathtt{pre}(a(\vec{o}))$. The resulting state $s'$ of applying $a(\vec{o})$ in $s$ is defined as follows: $s'=\mu(s, a(\vec{o}))=(s \backslash 
 \mathtt{del}(\vec{o}))\cup \mathtt{add}(\vec{o})$. 
A plan of a GP instance $Q$ is an action sequence $\alpha_n$ whose execution induces a state sequence $s_0,\cdots,s_n$ s.t. for each $1\leq i\leq n$, $a_i$ is applicable in $s_{i-1}$, and $s_i=\mu(s_{i-1}, a_i)$, and $s_n$ satisfies the goal of $Q$.

\begin{example}\rm \label{Gripper domain}
The GP problem $\mathcal{Q}^{l}_{grp}$ involves a robot with two grippers and several balls placed in different rooms. A gripper can carry only one ball at a time and the robot can move between rooms. Predicates are as follows: $gripper(g)$ means $g$ is a gripper; $ball(b)$ means $b$ is a ball;
$free(g)$ means $g$ is free; $carry(b,g)$ means gripper $g$ carries ball $b$; $at\textit{-}robby(r)$ means robby is at $r$; $at(b,r)$ means ball $b$ is at room $r$; $goal\textit{-}at(b,r)$ means ball $b$ need to be transported to goal room $r$.
Actions are as follows: $move(r, r')$ denotes robby moves from room $r$ to room $r'$; $drop(b,r,g)$ denotes $g$ drops $b$ at $r$; $pick(b,r,g)$ denotes $g$ picks $b$ at $r$. 
A GP problem is $\mathcal{Q}^{l}_{grp} = \langle \mathcal{D}, S_0,  S_G\rangle$, 
where initial condition $S_0$ is $\exists b,r,g.at(b,r)\land free(g)\land at\text{-}robby(r)$, and goal condition $S_G$ is $\forall b,r.at(b,r)$. 
An instance of $\mathcal{Q}^{l}_{grp}$ is $Q^{l}_{grp}=\langle \mathcal{D}, s_0,  s_g\rangle$, where the initial state is $s_0=\{\text{at}(b_1,r_1), \text{at}(b_2,r_2),  \text{at}(b_4,r_4), \text{free}(g_1), \allowbreak \text{free}(g_2), \text{at-robby}(r_1)\}$, and the goal condition is $s_g =\{\text{at}(b_1,r_3), \text{at}(b_2,r_5), \text{at}(b_4,r_1)\}$.

\end{example}

\subsection{Qualitative Numerical Planning}

Let $\mathcal{B}$ be a set of propositional variables and  $\mathcal{X}$ be a set of non-negative numerical variables. $\mathcal{L}$ denotes the class of all consistent sets of literals of the form $x>0$ and $x=0$ for $x\in \mathcal{X}$, $p$ and $\neg p$ for $p\in \mathcal{B}$. 

A QNP domain is a tuple $\mathcal{D}=\langle \mathcal{B}, \mathcal{X}, \mathcal{A}\rangle$, where $\mathcal{B}$ is a set of propositional variables, $\mathcal{X}$ is a set of non-negative numerical variables, $\mathcal{A}$ is a set of actions. Each $a \in \mathcal{A}$ has a set of preconditions $\mathtt{pre}(a) \in \mathcal{L}$, propositional effects \textit{eff}$(a) \in \mathcal{L}_B $ and numerical effects \textit{n-eff}$(a)$ that only contain special atoms of the form $ inc(v) $ or $ dec(v) $ to increase or decrease $v$ by some arbitrary amount for $ v \in V $. Actions with the $ dec(v) $ effect must feature the precondition $ v > 0 $ for any variable $ v \in V $.

A QNP problem is a tuple $\mathcal{Q} = \langle \mathcal{D}, S_0, S_G \rangle$, where $\mathcal{D}$ is a QNP domain, $S_0 \in \mathcal{L}$ denotes the initial condition, and $ S_G \in \mathcal{L}$ denotes the goal condition. A \emph{qualitative state} (qstate) $s$ of $\mathcal{Q}$ is an element of $ \mathcal{L}$ that contains a literal of each variable. 
An instance of $\mathcal{Q}$ is a \emph{numerical planning} problem $Q = \langle \mathcal{D}, s_0, s_g \rangle$ which replaces $S_0$ and $S_G$ with states $s_0$ and $s_g$ satisfying $S_0$ and $S_G$, respectively. A \emph{(quantitative) state} of $Q$ is an assignment of non-negative values to all $x \in \mathcal{X}$ and of truth values to $p \in \mathcal{B}$. For a state $\bar{s}$ and a qstate $s$, $\bar{s}$ satisfies $s$ if the assignment of values for variables in $\bar{s}$ satisfies the literals in $s$. 

A policy $\pi$ for a QNP problem is a partial mapping from qstates into actions. Given a policy $\pi$, a $\pi$-trajectory is a (finite or infinite) sequence of states $ \bar{s}_0, \bar{s}_1, \dots $ s.t. for $i \ge 0$,  $ \bar{s}_{i+1}$ can be resulted from executing $\pi(\bar{s}_i)$ where $s_i$ is the qstate satisfied by $s_i$. Given a QNP problem $\mathcal{Q}$, a policy $\pi$ solves an instance of $\mathcal{Q}$ if  any $\pi$-trajectory started at the initial state is goal reaching. A policy $ \pi $ solves $\mathcal{Q}$ if it solves every instance of $\mathcal{Q}$. For a given QNP problem $Q$, a policy $\pi$ terminates if for any instance of $Q$, every $\pi$-trajectory started at the initial state is finite. \citeauthor{ZengLL22} \shortcite{ZengLL22} developed a QNP solver, called DSET, based on AND/OR Graph Search.

\begin{example}\rm \label{abs-Gripper domain}
    The QNP abstraction $\mathcal{Q}^{h}_{grp}$ of Example \ref{Gripper domain} involves one numerical variable $N$ that denotes the number of balls that have not been transported to the target rooms. There are also three boolean variables $H, A$ and $G$, which denote the grippers carrying at least one ball, the robot being in the source room of a ball, and the robot being in the target room of a ball carried by grippers, respectively. Actions are as follows: $Move\textit{-}Ball$ means the robot moves to the room containing balls; $Pick$ means the robot picks up a ball; $Move\textit{-}Goal$ denotes the robot moves to a target room; $Drop$ denotes the robot drops a ball in the goal room.  
    The initial state of $\mathcal{Q}^{h}_{grp}$ is $S_0=\{N>0,\neg H,\neg A,\neg G\}$, and the goal condition of $\mathcal{Q}^{h}_{grp}$ is $S_G=\{N=0,\neg H,\neg A,\neg G\}$.
\end{example}







\subsection{Abstraction for Generalized Planning}
\citeauthor{2018Features} \shortcite{2018Features} proposed an abstraction framework using QNP for GP.  We adapt the framework to our setting.

We assume that there are two planning domains, one is an QNP domain $\mathcal{D}_{h}=\langle \mathcal{B}, \mathcal{X}, \mathcal{A}_h\rangle$ and the other is a GP domain $\mathcal{D}_{l}=\langle \mathcal{P}, \mathcal{O}, \mathcal{A}_l\rangle$. We consider $\mathcal{D}_h$ as a high-level (HL) domain and $\mathcal{D}_l$ a low-level (LL) domain. 
To define the QNP abstractions for GP problems, we first introduce the notion of refinement mapping $m$ to relate expressions between two languages of $\mathcal{D}_h$ and $\mathcal{D}_l$. 


\begin{definition}\rm 
We say that a function $m$ is a refinement mapping from $\mathcal{D}_{h}=\langle \mathcal{B}, \mathcal{X}, \mathcal{A}_h\rangle$ to $\mathcal{D}_{l}=\langle \mathcal{P}, \mathcal{O}, \mathcal{A}_l\rangle$ iff: for each HL propositional variable $p$, $m(p)=\phi_{p}$, where $\phi_{p}$ is a LL closed FOL formula; for each HL numerical variable $n$, $m(n)=\tau_n$, where $\tau_n$ is a LL counting term; for each HL action $a_h$, $m(a_h)=a_l(\vec{x})$, where $a_l(\vec{x}) \in \mathcal{A}_l$. 
\end{definition}
\begin{example}\rm \label{refinement-mapping-Gripper domain}
    The QNP problem $\mathcal{Q}^h_{grp}$ is exactly an abstraction of the GP problem $\mathcal{Q}^l_{grp}$, refinements of a propositional variable, a numerical variable and two actions of $\mathcal{Q}^h_{grp}$ are as follows:
    \begin{itemize}[leftmargin=*]
        \item $m(Pick)=pick(b,r,g)$;
        \item $m(Drop)=drop(b,r,g)$;
        \item $m(H) = \exists b,g,r. (\text{carry}(b,g) \land \text{goal\_at}(b,r))$;
        \item $m(N)=\# b. ((\exists rg. \text{goal\_at}(b,rg)$
        
        \hspace{5.5em}$\land \neg\exists r.(\text{at}(b,r) \land \text{goal\_at}(b,r)))$.
    \end{itemize}
\end{example}

Let $\mathcal{Q}_h$ be a QNP problem and $\mathcal{Q}_l$ be a GP problem,  we hereafter introduce the concept of $m$-isomorphism between HL and LL states. In the following, we assume that the HL and LL domains share the same variable assignment $v$. 
	
\begin{definition}\rm \label{refinement-mapping}
		Given a refinement mapping $m$, a HL state $s_{h}$ is $m$-isomorphic to a LL state $s_{l}$, written $s_{h}\sim_{m}s_{l}$, if: for any HL propositional variable $p$, we have $s_h\models p$ iff $s_l\models m(p)$; for any HL numerical variable $n$, we have $s_h\models n=y$ iff $s_l\models m(n)=y$
\end{definition}




We now present a notion of $m$-simulation relation between a HL instance $Q_h$ and a LL instance $Q_l$. $\Delta^Q_S$ denotes the set of all possible states in $Q$.

\begin{definition}\rm \label{abs-relation} 
 A relation $R\subseteq \Delta^{Q_h}_S\times \Delta^{Q_l}_S$ is called an $m$-simulation relation between $Q_h$ and $Q_l$, iff $\langle s^h_0, s^l_0 \rangle \in R$, and $\langle s_h, s_l \rangle \in R$ implies that:  $s_{h}\sim_{m}s_{l}$; for any HL action $a$, if there is a state $s'_{h}$ in $Q_h$, s.t. $s'_h\in \eta(s_h, a)$, then there is a state $s'_{l}$ in $Q_l$, s.t. $s'_l=\mu(s_l, m(a))$ and $\langle s'_{h},s'_{l}\rangle \in R$.
\end{definition}

\begin{definition}\rm \label{Sound}
    Let $\mathcal{Q}_h$ be a QNP problem and $\mathcal{Q}_l$ a GP problem, $\mathcal{Q}_h$ is called a sound abstraction of $\mathcal{Q}_l$ if for each instance $Q_l$ in $\mathcal{Q}$, there is an instance $Q_h$ of $\mathcal{Q}_h$ s.t. there is a $m$-simulation relation $R$ between $Q_h$ and $Q_l$, and for any $\langle s_h, s_l \rangle \in R$, $s_h\models s^h_g$ iff $s_l\models s^l_g$.
\end{definition}

\begin{theorem}
    If a QNP problem $\mathcal{Q}_h$ is a sound abstraction of a GP problem $\mathcal{Q}_l$ and a policy $\pi$ is a solution of $\mathcal{Q}_h$, then $m(\pi)$ is a solution of $\mathcal{Q}_l$.
\end{theorem}

\begin{definition}\rm \label{A-Sound}
   Let $\mathcal{Q}_h$ be a QNP problem and $\mathcal{Q}_l$ a GP problem, $\mathcal{Q}_h$ is called an approximate sound abstraction relative to a finite set $\mathcal{I}$ of instances of $\mathcal{Q}_l$ if $\mathcal{Q}_h$ is a sound abstraction of $\mathcal{I}$.
\end{definition}

\begin{corollary}\label{hl-plan-2-LL-plan}
      If a QNP problem $\mathcal{Q}_h$ is an approximate sound abstraction relative to a finite instance set $\mathcal{S}_l$ of a GP problem $\mathcal{Q}_l$ and $\pi$ is a solution of $\mathcal{Q}_h$, then $m(\pi)$ is a solution of $\mathcal{S}_l$.
\end{corollary}

\section{QNP Abstraction Generation}

\subsection{Prompt Protocol}

We are interested in whether LLMs can be used to generate QNP abstractions for GP problems. 
To this end, we first define the prompt protocols.

\noindent\textbf{Feature Generation.} We first ask LLM to generate features given an input GP domain:

\begin{mdframed}[
  linecolor=black, 
  linewidth=1pt,  
  backgroundcolor=gray!10,
  innerleftmargin=5pt,   
  innerrightmargin=5pt,   
  skipabove=10pt,     
  skipbelow=10pt,     
  align=left         
]
\begin{tabular}{@{}p{\dimexpr\linewidth-5pt\relax}@{}} 
Input: Domain $\mathcal{D}_l$\\
--\ --\ --\ --\ --\ --\ --\ --\ --\ --\ --\ --\ --\ --\ --\ --\ --\ --\ --\ --\ --\ --\ --\ --\ --\\
Generate Boolean and numerical features according to the following template,
\begin{itemize}[leftmargin=*]
    \item Boolean feature: $p \defeq \exists \vec{x}.\ \phi(\vec{x})$, where $\phi(\vec{x})$ is a first-order formula over $\mathcal{D}_l$.
    \item Numerical feature:  $n \defeq \# \vec{x}. \delta(\vec{x})$, where $\delta(\vec{x})$ is a first-order formula over $\mathcal{D}_l$.  
\end{itemize}
Output the Boolean feature set $\mathcal{B}$ and numerical feature set $\mathcal{X}$.
\end{tabular}
\end{mdframed}

\noindent\textbf{Abstraction Generation.} We then ask for a QNP abstraction given GP instances (or LL instances):

\begin{mdframed}[
  linecolor=black, 
  linewidth=1pt,  
  backgroundcolor=gray!10,
  innerleftmargin=5pt,   
  innerrightmargin=5pt,   
  skipabove=10pt,     
  skipbelow=10pt,     
  align=left         
]
\begin{tabular}{@{}p{\dimexpr\linewidth-5pt\relax}@{}} 
Input: Domain $\mathcal{D}_l$, Instances $\{Q_l^1,\ldots Q_l^n\}$\\
--\ --\ --\ --\ --\ --\ --\ --\ --\ --\ --\ --\ --\ --\ --\ --\ --\ --\ --\ --\ --\ --\ --\ --\ --\\
Generate a QNP abstraction for instances based on the feature as follows.
\begin{itemize}[leftmargin=*]
    \item Step 1: compute the abstraction $S_0$ for the initial states in $\{Q_l^1,\ldots Q_l^n\}$;
    \item Step 2: compute the abstraction $S_G$ for the goals in $\{Q_l^1,\ldots Q_l^n\}$; 
    \item Step 3: compute the abstraction action set $\mathcal{A}_h$ based on the action set $\mathcal{A}_l$ of domain $\mathcal{D}_l$
\end{itemize}
Output QNP  $\mathcal{Q}=\langle \mathcal{D}_h, S_0, S_G\rangle$, where $\mathcal{D}_h=\langle \mathcal{B}, \mathcal{X}, \mathcal{A}_h\rangle$, and refinement mapping $m$.
\end{tabular}
\end{mdframed}

Example features generated by GPT-5.2 are shown in Example \ref{refinement-mapping-Gripper domain}, and an QNP abstraction generated by GPT-5.2 are shown in Example \ref{abs-Gripper domain}.

\subsection{Automated Interactive Debugging}


For the QNP abstraction $\mathcal{Q}=\langle \mathcal{D}, S_0, S_G \rangle$ provided by LLM, we examine the approximate soundness of $\mathcal{Q}$ with the given LL training instances. We perform 4 steps: 
(1) Check whether $\mathcal{Q}$ has a solution policy $\pi$; 
(2) For each training instance $Q_l$, check whether $\pi$ can be executed as a plan $\sigma_h$  to solve the HL instance $Q_h$ induced by refinement mapping $m$ with $Q_l$; 
(3) Check whether the plan $\sigma_h$ can be refined and executed in $Q_l$; 
(4) Check whether the refined plan for $Q_l$ can reach the goal. For each step, if the abstraction can pass, we proceed to the next step; otherwise, we report errors to LLM and prompt it to fix the abstraction. We set an upper bound N on the number of prompts allowed for debugging. If the number of prompts exceeds the upper bound, we consider that the LLM cannot provide an useful abstraction. The following is the detailed process:

\paragraph{Abstraction Solvability Check (ASC).} For the QNP abstraction $\mathcal{Q}$, we first leverage DSET \cite{ZengLL22} to solve it: if DSET returns that $\mathcal{Q}$ has no solution, or if DSET encounters an unexpected dead-end leading to a timeout, we feedback the corresponding information to LLM and prompt it to fix the abstraction; if DSET returns a policy $\pi$ for $\mathcal{Q}$, we proceed to the subsequent checking step. An example prompt template is shown below.
\vspace{-1mm}
\begin{mdframed}[
  linecolor=black, 
  linewidth=1pt,  
  backgroundcolor=gray!10,
  innerleftmargin=5pt,   
  innerrightmargin=5pt,   
  skipabove=8pt,     
  skipbelow=8pt,     
  align=left         
]
\begin{tabular}{@{}p{\dimexpr\linewidth-5pt\relax}@{}} 
Input: QNP Abstraction $\mathcal{Q}$\\
--\ --\ --\ --\ --\ --\ --\ --\ --\ --\ --\ --\ --\ --\ --\ --\ --\ --\ --\ --\ --\ --\ --\ --\ --\\
We used a QNP solver to solve $\mathcal{Q}$, but no solution is return. Fix the QNP abstraction $\mathcal{Q}$.
\end{tabular}
\end{mdframed}

\paragraph{HL Instance Solvability Check (HLISC).} 
We first examine whether the QNP abstraction $\mathcal{Q}$ is suitable for 
LL instance $Q_l$, i.e., whether there is a HL instance $Q_h$ of $\mathcal{Q}$
for $Q_l$. To do so, we compute the HL instance $Q_h$ of $Q_l$ using 
the refinement mapping $m$. We obtain $Q_h$ which is composed of the abstract state $s^h_0$ of the initial state $s^l_0$ in $Q_l$, the abstract state $s^h_g$ of the goal $s^l_g$ in $Q_l$ and the domain of $\mathcal{Q}$. 
We check whether $s^h_0$ (resp. $s^h_g$) satisfies the initial condition $S_0$ (resp. goal $S_G$) of $\mathcal{Q}$. If so, $Q_h$ is confirmed as an instance of $\mathcal{Q}$; otherwise, the abstract features generated by the LLM are inappropriate, then we prompt LLM to fix the abstraction.


We then iteratively execute the policy $\pi$ starting from $s^h_0$ of $Q_h$. During execution, the following two exceptions may occur: (a) The execution of $\pi$ aborted prematurely before reaching the goal of $Q_h$. (b) The execution of $\pi$ timed out due to getting stuck in an unexpected dead-end. In each case, we report the corresponding information to LLM and prompt it to fix the abstraction. If neither of these two cases occur, we output the action sequence $\sigma_h$ induced by the policy $\pi$, which is also a valid plan for $Q_h$. An example prompt template is shown below.

\begin{mdframed}[
  linecolor=black, 
  linewidth=1pt,  
  backgroundcolor=gray!10,
  innerleftmargin=5pt,   
  innerrightmargin=5pt,   
  skipabove=10pt,     
  skipbelow=10pt,     
  align=left         
]
\begin{tabular}{@{}p{\dimexpr\linewidth-5pt\relax}@{}} 
Input: QNP $\mathcal{Q}$, LL $Q_l$, Policy $\pi$, Mapping $m$\\
--\ --\ --\ --\ --\ --\ --\ --\ --\ --\ --\ --\ --\ --\ --\ --\ --\ --\ --\ --\ --\ --\ --\ --\ --\\
We compute $Q_h$ with $Q_l$ and $m$, and confirmed that $Q_h$ is an instance of $\mathcal{Q}$. Policy $\pi$ should solve $Q_h$, but fails.
Exception:
The execution $exc(\pi)$ of $\pi$ aborted prematurely before reaching the goal of $Q_h$.
Fix the QNP abstraction $\mathcal{Q}$.
\end{tabular}
\end{mdframed}

\begin{algorithm}[tb]
\caption{Refined Tree Construction} 
\label{alg:refine-tree}

\KwIn{$Q_l$, $m$, $Q_h$, $\sigma_h$}
\KwOut{Success / Error / Refined tree T} 
$T \gets \emptyset$, $x\gets 0$, $s_l\gets s^l_0$, $s_h\gets s^h_0$, 
$k \gets \lvert \sigma_h \rvert$\;

\If{\textnormal{\textbf{DFS}}($T,\sigma_h,s_h,s_l,x$) is Success}{
  \Return Success\;
}
\lIf{$depth(T)< k$}{ \Return Error}
\lIf{$depth(T)\geq k$}{ \Return T}  
\smallskip
where \textbf{DFS} is the following procedure:\\
\smallskip
    \If{$x \geq k$, $s_h\models s^h_g$ and $s_l\models s^l_g$}{
       \Return Success\;
    }  
    $a_h \leftarrow action(\sigma_h, s_h)$, $A_l \leftarrow applicable(s_l)$\;
    \For{each $a_l \in A_l$}{
        \If{$a_l$ is not a refinement of $a_h$}{
            \textbf{continue}\;
        }
        $s_l' \gets next(s_l, a_l)$, $s_h' \gets next(s_h, a_h)$ \;
        $\tilde{s}_h' \gets compute\text{-}abstraction(m, s_h)$\;
        \lIf{$s_h' \neq \tilde{s}_h'$}{
            \textbf{continue}
        }
        $\text{child} \gets \text{Expand}(T, s_l', s_h', a_l, a_h, x+1)$\;
        $T.current \gets \text{child}$\;
        \textbf{DFS}{(T, $\sigma_h, s_h', s_l', x+1$)}\;
        $T.current \gets T.current.parent$\;
    }
\end{algorithm}

\paragraph{HL Plan Refinability Check (HLPRC).}
To check the approximate soundness of $\mathcal{Q}$, 
we refine the HL plan $\sigma_h$ to iteratively construct a part of a  $m\textit{-}$simulation relation between $Q_h$ and $Q_l$ (Algorithm 1), and identify potential abstraction errors based on LL information.

The inputs of Algorithm 1 are a LL instance $Q_l$, HL instance $Q_h$, a plan $\sigma_h$ for $Q_h$, and refinement mapping $m$; The output is $Success$, $Error$ or a state transition tree (or refined tree) $T$ of $Q_l$.
In Algorithm 1, guided by the HL plan $\sigma_h$, we start from the initial state $s^l_0$ of $Q_l$ and expand its next states in a depth-first manner to construct a state-transition tree, which we called a refined tree $T$ of $Q_l$. 
Specifically: First, we set $s^l_0$ as the root of $T$ with layer 0. Since $s^h_0$ is an abstraction of $s^l_0$, $s^h_0$ and $s^l_0$ are $m$-isomorphic. Then, for any non-goal state $s^h_i$ reachable from $s^h_0$ and a non-goal state $s^l_i$ with layer $i$ reachable from $s^l_0$, assuming that $s^h_i$ and $s^l_i$ are $m$-isomorphic, we now have the following 2 cases for the successor actions and states:


\textbf{Case 1:} For the HL action $a^h_i$ in $s^h_i$, if there exists no LL action $a^l_i$ that is applicable in $s^l_i$ such that $a^l_i$ is a refinement of $a^h_i$, then the algorithm records the HL and LL actions, backtracks to the parent of $s^l_i$ and continues to expand.  



\textbf{Case 2:} There exists an action $a^h_i$ applicable in state $s^h_i$ and an action $a^l_i$  applicable in state $s^l_i$ such that $a^l_i$ is a refinement of $a^h_i$. Let $s^h_{i+1}$ be a successor state of $s^h_i$, the algorithm expands $s^l_i$ to a new state $s^l_{i+1}$ with layer $i+1$ based on action $a^l_i$, and then check whether $s^h_{i+1}$ is $m$-isomorphic to $s^l_{i+1}$. If this is not the case, the algorithm records the HL and LL transitions $(s_i,a_i,s_{i+1})$ and continues to expand; otherwise, we have the following 3 subcases: (a) $s^h_{i+1} \nvDash s^h_g$ and $s^l_{i+1}\nvDash s^l_g$, for this case, the algorithm continues to expand; (b) $s^h_{i+1}\models s^h_g$ and $s^l_{i+1} \models s^l_g$, for this case, the algorithm returns $Success$; (c) $s^h_{i+1} \models s^h_g$ and $s^l_{i+1} \nvDash s^l_g$, for this case,  the algorithm continues to expand.
Note that the subcase where $s^h_{i+1}\nvDash s^h_g$ and $s^l_{i+1}\models s^l_g$ is impossible, since $s^h_{i+1}$ and $s^l_{i+1}$ are $m$-isomorphic. An example prompt template is shown below.

\vspace{-1.5mm}
\begin{mdframed}[
  linecolor=black, 
  linewidth=1pt,  
  backgroundcolor=gray!10,
  innerleftmargin=5pt,   
  innerrightmargin=5pt,   
  skipabove=10pt,     
  skipbelow=10pt,     
  align=left         
]
\begin{tabular}{@{}p{\dimexpr\linewidth-5pt\relax}@{}} 
Input: Instances $Q_l$ and $Q_h$, Plan $\sigma_h$ of $Q_h$\\
--\ --\ --\ --\ --\ --\ --\ --\ --\ --\ --\ --\ --\ --\ --\ --\ --\ --\ --\ --\ --\ --\ --\ --\ --\\

Plan $\sigma_h$ should be refined as an action sequence in $Q_l$, when we trying to do this, the following exception raised:
\begin{itemize}[leftmargin=*]
    \item There is no refinement action in $Q_l$ of the abstract action $a_h$. Action $a_h$ should be an abstraction of one of actions in $\{a^l_1, ...a^l_n\}.$
\end{itemize}
Fix the QNP abstraction $\mathcal{Q}$.




\end{tabular}
\end{mdframed}

If no goal state is found after the tree has been fully expanded to depth \(k\), where \(k = \lvert \sigma_h \rvert\), we return the tree.
In the next step, we use the LL states and actions appearing in the returned tree to localize potential abstraction errors.

\paragraph{LL Goal Reachability Check (LLGRC).} 
Given a refined tree $T$, we diagnose the potential errors appear in the abstraction based on Definition 4.
Excluding the nodes with layer $k$, we perform a $(k-i)$-step goal reachability check\footnote{The checking is performed by Fast Downward.} 
in a bottom-up manner on the nodes at level $i$ of $T$. For any node $s^l_i$ with layer $i$ of $T$,  if $s^l_i$ is $(k-i)$-step goal-reachable, and all the successor nodes $s^l_{i+1}$ of $s^l_i$ are $(k-i-1)$-step goal-unreachable, then it indicates that the HL transition $(s^h_i, a^h_i, s^h_{i+1})$ is inappropriate. An example prompt template\footnote{All notations in the prompt template will be replaced with their corresponding files in experiments.} is shown below.

\begin{mdframed}[
  linecolor=black, 
  linewidth=1pt,  
  backgroundcolor=gray!10,
  innerleftmargin=5pt,   
  innerrightmargin=5pt,   
  skipabove=10pt,     
  skipbelow=10pt,     
  align=left         
]
\begin{tabular}{@{}p{\dimexpr\linewidth-5pt\relax}@{}} 
Input: Instances $Q_l$ and $Q_h$, Plan $\sigma_h$ of $Q_h$, Action sequence $\sigma_l$ in $Q_l$\\
--\ --\ --\ --\ --\ --\ --\ --\ --\ --\ --\ --\ --\ --\ --\ --\ --\ --\ --\ --\ --\ --\ --\ --\ --\\
The action sequence $\sigma_l$ refined by $\sigma_h$ should reach the goal of $Q_l$. When we examined the request, the following exception raised:
\begin{itemize}[leftmargin=*]
    \item The abstract action $a_h$ and state $s_h$ in $Q_h$ corresponds to $a_l$ and $s_l$ in $Q_l$ are inappropriate.
\end{itemize}
Fix the QNP abstraction $\mathcal{Q}$.
\end{tabular}
\end{mdframed}

\section{Experiments and Results}

\begin{table*}[t]
\centering
\normalsize 
\begin{tabular*}{\textwidth}{@{\extracolsep{\fill}}l *{8}{S[table-format=1.2]}}
\toprule
\multirow{2}{*}{\textbf{Domain}} & \multicolumn{2}{c}{\textbf{GPT}} & \multicolumn{2}{c}{\textbf{Gemini}} & \multicolumn{2}{c}{\textbf{DeepSeek}} & \multicolumn{2}{c}{\textbf{Qwen}} \\
\cmidrule(lr){2-3} \cmidrule(lr){4-5} \cmidrule(lr){6-7} \cmidrule(lr){8-9}
 & {\makebox[0pt][c]{\ding{52}AD}} & {\makebox[0pt][c]{\ding{56}AD}} & {\makebox[0pt][c]{\ding{52}AD}} & {\makebox[0pt][c]{\ding{56}AD}} & {\makebox[0pt][c]{\ding{52}AD}} & {\makebox[0pt][c]{\ding{56}AD}} & {\makebox[0pt][c]{\ding{52}AD}} & {\makebox[0pt][c]{\ding{56}AD}} \\
\midrule
Delivery & 1.00 & 0.70 & 0.50 & 0.50 & 0.50 & 0.00 & 0.00 & 0.00 \\
Heavy    & 1.00 & 0.75 & 1.00 & 1.00 & 0.50 & 0.00 & 0.50 & 0.00 \\
Gripper  & 0.85 & 0.00 & 0.30 & 0.00 & 0.05 & 0.00 & 0.00 & 0.00 \\
Miconic  & 0.40 & 0.00 & 0.00 & 0.00 & 0.00 & 0.00 & 0.00 & 0.00 \\
Ferry    & 1.00 & 0.15 & 1.00 & 0.15 & 0.55 & 0.00 & 0.00 & 0.00 \\
Spanner  & 0.00 & 0.00 & 0.00 & 0.00 & 0.00 & 0.00 & 0.00 & 0.00 \\
Forest   & 0.00 & 0.00 & 0.00 & 0.00 & 0.00 & 0.00 & 0.00 & 0.00 \\
\bottomrule
\end{tabular*}
\small
\vspace*{-2mm}
\caption{The coverage rate (in [0,1]) of QNP abstraction generated by LLMs on the evaluation set. Results are averaged over 4 runs and 10 evaluation instances per run. AD is the abbreviation of automated debugging.}
\label{tab:debug_performance}
\end{table*}

\begin{table}[t]
\centering
\normalsize  
\setlength{\tabcolsep}{3pt} 
\begin{tabular*}{\columnwidth}{@{\extracolsep{\fill}}l S[table-format=2.3] S[table-format=2.2] S[table-format=2.2] S[table-format=2.2]}
\toprule
\textbf{Error Source} & \multicolumn{1}{c}{\textbf{GPT}} & \multicolumn{1}{c}{\textbf{Gemini}} & \multicolumn{1}{c}{\textbf{DeepSeek}} & \multicolumn{1}{c}{\textbf{Qwen}} \\
\midrule
ASC & 2.93 & 4.50 & 6.11 & 8.36 \\
HLISC & 0.46 & 0.46 & 1.04 & 0.50 \\
HLPRC & 1.64 & 1.61 & 1.82 & 0.43 \\
LLGRC & 0.00 & 0.00 & 0.00 & 0.00 \\
\bottomrule
\end{tabular*}
\caption{Average number of detected errors (per run)}
\vspace*{-3mm}
\label{tab:table2}
\end{table}



\paragraph{Experimental Setup.}

We conduct experiments on 7 domains from the generalized planning literature~\cite{SilverDSTK024}, we now briefly describe these domains: 
\textbf{Delivery}: Newspapers at a home base must be delivered to multiple locations; \textbf{Forest}: A hiker must navigate a 2D grid to reach a goal location while climbing hills and avoiding water; \textbf{Gripper}: Balls must be transported between rooms by a robot with two grippers; \textbf{Miconic}: Passengers in multiple buildings, each with an elevator, must be picked up and dropped off on different floors; \textbf{Ferry}: Cars must be sailed between islands using a ferry that can carry at most one car; \textbf{Spanner}: Spanners and nuts are distributed along a one-way corridor. An agent moves down the corridor, pick up wrenches, and tighten the nuts, using each wrench at most once; \textbf{ Heavy}: Items must be stacked in an empty box. An item can only be stacked on another item if the latter is heavier. 

For each domain, we use 4 training instances to generate a QNP abstraction and 10 evaluation instances to assess the resulting abstraction.
All instances are randomly generated. Each instance contains 10 to 30 objects.
Both domain descriptions and problem instances are provided to the LLMs as PDDL files. Our experiments were run on a Windows 11 machine with a 2.60 GHz Intel Core i5-13500H CPU and 16.0 GB RAM.

We evaluated 4 state-of-the-art LLMs for generating QNP abstractions: GPT-5.2 (Extended Thinking), Gemini-3 (Pro), DeepSeek-V3.2 (Reasoner), and Qwen-3 (Max Thinking). For each LLM, we tested two settings: direct generation (no automated debugging) and generation with automated debugging. 
In the no-debug setting, 4 training instances are provided and the LLM generates the QNP abstraction in a single pass. In the debug setting, 2 training instances are used for initial generation, and the remaining 2 training instances are provided jointly for debugging. 
To account for the stochasticity of LLM decoding, we repeat each setting 4 times using the same training and evaluation instances and report the mean number of 
solved evaluation instances across the 4 runs. In each run, a single QNP abstraction is evaluated on 10 evaluation instances. In the debugging setting, we allow at most 10 debugging iterations per run.

\paragraph{Results and Analysis.} Table 1 shows that regardless of whether automated debugging is applied, QNP abstractions generated by GPT cover the broadest range of GP instances, demonstrating its powerful inference and generalization capabilities. Gemini's capabilities are weaker than those of GPT. Qwen fails to generate useful QNP abstractions in most cases.  Experimental results indicate that, except for Qwen, our automated debugging method generally improves performance dramatically. 

\paragraph{Error source:} For all LLMs, errors in abstraction are mainly fixed by ASC followed by HLPRC. 
We did not encounter any errors that required LLGRC on our tested domains.
However, this does not imply that LLGRC is unnecessary: according to the definition of approximate soundness (Definition~4), the LLGRC check is still required in general.

\paragraph{Domain failure:} In Spanner domain, GPT consistently provides an abstract action $walk\text{-}to\text{-}spanner$ to characterize the agent moving from a position without a spanner to one with a spanner. The refinement of the action should be an LL action sequence, e.g. $walk(r0, r1), walk(r1, r2)$, which is not covered by our formalization. In addition, for the one-way predicate $link$ connecting positions 1 and 2, GPT often assumes that these two positions are commutative. In Forest domain, a key numerical feature should be the distance between two coordinates, but the underlying logic FOC used in our work cannot express these features.




\section{Discussion and Future Work}

We have shown that some LLMs, especially GPT-5.2, can serve as useful QNP abstraction generators for GP when coupled with our carefully designed automated debugging procedure. The following are limitations. First, the expressive ability of predefined feature templates is relatively limited, and the refinement of HL actions must be an LL action rather than an action sequence or a program, which narrows the applicable scope of our method. Second, the information why the abstraction fails, provided by our method, is not sufficiently precise, which may affect the accuracy of generated abstractions. Third, our work can not guarantee the correctness of the generated abstractions. These limitations motivate several directions for future work.

\appendix


\bibliographystyle{named}
\bibliography{ijcai26}

\end{document}